\documentclass{IEEEtran}

\usepackage[activate={true,nocompatibility},final,tracking=true,kerning=true,spacing=true,factor=1100,stretch=10,shrink=10]{microtype}  % Reduce page count
\usepackage{parskip}  % Disables indenting
\usepackage{nth}
\usepackage{float}
\usepackage[font=small,labelfont=bf]{caption}  % Subfigures
\usepackage[font=small,labelfont=bf]{subcaption} % Subfigures
\usepackage[lmargin=25mm,rmargin=25mm,tmargin=27mm,bmargin=30mm]{geometry}
\usepackage{graphicx}
\usepackage{upgreek}
\usepackage{amssymb}
\usepackage{mathrsfs}
\usepackage{amsmath}
\usepackage{color}
\usepackage{csquotes}
\usepackage{tikz}
\usepackage{bm}  % Bold math symbols
\usepackage{multirow}  % Table cells that span multiple rows and cols
\usepackage{bigstrut}  % \bigstrut gives extra space in table rows
\usepackage{listings}  % Used to display syntax highlighted code
\usepackage[disable]{todonotes}
\usepackage{enumitem} % Easily customisable enumerate symbols
\usepackage[ruled,vlined]{algorithm2e}  % Algorithms with style
\usepackage[pdftex]{hyperref}
\let\oldbibliography\bibliography
\renewcommand{\bibliography}[1]{%
  \oldbibliography{#1}%
  \setlength{\itemsep}{0pt}%
}

\newcommand{\bmr}[1]{\bm{\mathrm{#1}}}
\newcommand{\bmat}[1]{\begin{bmatrix}#1\end{bmatrix}} 
\newcommand{\mat}[1]{\begin{matrix}#1\end{matrix}} 
\DeclareMathOperator{\sign}{sgn}

\newcommand{\tikzdot}[2][red,fill=red]{\tikz[baseline=-0.5ex]\draw[#1,radius=#2] (0,0) circle ;}%

\usepackage{xcolor}
 
\definecolor{codegreen}{rgb}{0,0.6,0}
\definecolor{codegray}{rgb}{0.5,0.5,0.5}
\definecolor{codepurple}{rgb}{0.58,0,0.82}
\definecolor{backcolour}{rgb}{0.98,0.98,0.98}
 
\lstdefinestyle{mystyle}{
    backgroundcolor=\color{backcolour},   
    commentstyle=\color{codegreen},
    keywordstyle=\color{magenta},
    numberstyle=\tiny\color{codegray},
    stringstyle=\color{codepurple},
    basicstyle=\ttfamily\footnotesize,
    breakatwhitespace=false,         
    breaklines=true,                 
    captionpos=b,                    
    keepspaces=true,                 
    showspaces=false,                
    showstringspaces=false,
    showtabs=false,                  
    tabsize=2,
    mathescape=true
}
 
\bstctlcite{IEEEexample:BSTcontrol}

\newcommand{\supercopyright}{\textsuperscript{\copyright}}

\usepackage[some,bottom]{background}
\SetBgContents{\parbox{1.05\textwidth}{\copyright 2019 IEEE. Personal use of this material is permitted. Permission from IEEE must be obtained for all other uses, in any current or future media, including reprinting/republishing this material for advertising or promotional purposes, creating new collective works, for resale or redistribution to servers or lists, or reuse of any copyrighted component of this work in other works.}}
\SetBgScale{0.9}
\SetBgOpacity{0.6}
\SetBgColor{black}
\SetBgVshift{1cm}

\begin{document}
\title{Dissecting Deep Neural Networks}
\author{Haakon Robinson$^{1}$, Adil Rasheed$^{1,2}$, Omer San$^{3}$
\thanks{$^{1}$Department of Engineering Cybernetics, Norwegian University of Science and Technology, Trondheim, Norway (e-mail: haakon.robinson@ntnu.no, adil.rasheed@ntnu.no)}% <-this % stops a space
\thanks{$^{2}$Department of Mathematics and Cybernetics, SINTEF Digital, Trondheim, Norway}
\thanks{$^{3}$Mechanical and Aerospace Engineering, Oklahoma State University, Stillwater, Oklahoma, USA (e-mail: osan@okstate.edu)}% <-this % stops a space
}
\maketitle

\BgThispage

\begin{abstract}
    In exchange for large quantities of data and processing power, deep neural networks have yielded models that provide state of the art predication capabilities in many fields. However, a lack of strong guarantees on their behaviour have raised concerns over their use in safety-critical applications. A first step to understanding these networks is to develop alternate representations that allow for further analysis. It has been shown that neural networks with piecewise affine activation functions are themselves piecewise affine, with their domains consisting of a vast number of linear regions. So far, the research on this topic has focused on counting the number of linear regions, rather than obtaining explicit piecewise affine representations. This work presents a novel algorithm that can compute the piecewise affine form of any fully connected neural network with rectified linear unit activations.
\end{abstract}

%%%%%%%%%%%%%%%%%%%%%%%%%%%%%%%%%%%%%%%%%%%%%%%%%%%%%%%%%%%%%%%%%%%%%%%%%%%%%%%%%%%%%
\section{Introduction}
    The recent successes of Machine Learning (ML) in image classification \cite{He2015}, \cite{inception-4}, \cite{NIPS2012_4824}) and games like GO can be largely attributed to Deep Neural Networks (DNN) \cite{Goodfellow-et-al-2016}. In particular, the ability to train extremely deep neural networks has yielded unprecedented performance in a myriad of fields. Examples of applications include diabetes detection \cite{KANNADASAN2018}, action detection for surveillance \cite{ULLAH2019386}, feature learning for process pattern recognition \cite{YU20191}, denoising for speech enhancement \cite{LIU2018106}, fault diagnosis \cite{SHAO2017187}, social image understanding \cite{LIU2019}, and low light image enhancements \cite{LORE2017650}.
    
    However, DNNs are still regarded as "black box" models, and few guarantees can be made about their behaviour. The idea of adversarial attacks have exposed that many existing DNNs models have very low robustness. It has been shown that by changing the input minimally in a targeted way, DNNs can be tricked into giving a completely wrong output. Such attacks are sometimes limited to a single pixel \cite{pixel-attacks}. This is especially problematic when applying DNN in safety critical applications like robotic surgery or autonomous cars. Therefore, the focus has now shifted to developing new methods for analyzing DNNs. 
    
    It has been shown that neural networks that only use piecewise affine (PWA) activation functions can themselves be expressed as a PWA function defined on convex polyhedra, although the number of regions can be enormous \cite{DBLP:journals/corr/EldanS15, montufar2017notes}. A succint description of the structure and combinatorics of PWA neural networks can be found in chapter 7 of \cite{strang2019linear}. A function is PWA if it can be defined piecewise over a set of polyhedral regions $\Omega_i$:
    
    \begin{equation}
        \label{eq:pwa}
        f(\bm{x}) = \bm{w}^T_i\bm{x} + \bm{b}_i \;\;\;\; \forall \bm{x} \in \Omega_i
    \end{equation}
    
    Methods for counting the number of regions have been developed, but little research has been done into explicitly finding and working with the PWA representation of neural networks, likely due to the high complexity of such a function. This is unfortunate, as there is a wealth of literature on PWA functions, particularly in the context of modeling and control. For example, it is well known that the explicit solution to the linear Model Predictive Control (MPC) problem is a PWA function, and schemes for using PWA models in the optimisation loop exist \cite{cervantes2003nonlinear}. Furthermore, there exist methods for verifying the stability of PWA systems, as well as stabilising them \cite{stability-pwa}. Positive invariant sets can be constructed for PWA systems by analysing the possible transitions between the linear regions of the system \cite{invariant-sets-PWA}. Thus, by decomposing a DNN into its PWA representation, these established methods can be used to obtain concrete stability results for a large and useful family of neural networks. Studies of the linear regions of neural networks started with the need to understand how \textit{expressive}\footnote{A more expressive network has the ability to compute more complex, rich functions.} these networks are, and how this changes with the architecture (number of layers, width of layers, etc) \cite{montufar2014number, pascanu2013number}. Expressivity is often measured using the Vapnik–Chervonenkis (VC) dimension \cite{vapnikchervonenkis}, and tight bounds have been found for the VC dimension of PWA neural networks \cite{bartlett2019nearly}. Empirical measures for the expressivity of PWA networks have also been developed \cite{Raghu:2017:EPD:3305890.3305975}.  Empirical evidence strongly suggests increasing the depth of a network has a bigger impact on expressivity than increasing the width of existing layers \cite{DBLP:journals/corr/EldanS15, DBLP:journals/corr/Telgarsky15}. In \cite{serra2018bounding} the authors present upper and lower bounds on the maximum number of regions that improve on previous results, along with a mixed-integer formulation from which the regions can be counted by enumerating the integer solutions. They established that for a network with input dimension $d$, number of hidden layers $L$ , each with $n$ nodes and ReLU activation, the asymptotic bounds for the maximal number of regions are:
    
    \begin{gather}
    \label{eq:bounds_regs}
    \begin{aligned}
        \text{Lower:}&\;\;\Omega\Big( (\tfrac{n}{d})^{(L-1)d}n^d \Big) \\
        \text{Upper:}&\;\;\mathcal{O}(n^{dL})
    \end{aligned}
    \end{gather}
    
    This upper bound is exponential in both $d$ and $L$. The most challenging aspect in terms of analysis is that the most useful neural networks are those with both large input dimension $d$ and large number of hidden layers $L$. The number of linear regions of such a network is enormous. It is likely due to this that there have been a limited number of studies into the identification of these regions. There have been studies on approximating nonlinear neural networks with PWA functions \cite{amin1997piecewise}. Conversely, work has been done on the inverse problem of representing PWA functions more compactly as neural networks \cite{wang2008configuration}.

    To this end, the main contribution of this work is an algorithm that can convert any neural network using fully connected layers and ReLU activations into its exact PWA representation that can be visualised and analysed, giving an insight into the inner workings of the network. This was achieved by utilising existing linear programming (LP) methods (specifically the MPT toolbox for MATLAB\supercopyright \cite{MPT3}) for working with polyhedral sets and hyperplane arrangements. The approach can also be extended to any linear / affine layer (convolutional layers, batch normalisation), as well as any PWA activation function (Leaky ReLU, maxout).

%%%%%%%%%%%%%%%%%%%%%%%%%%%%%%%%%%%%%%%%%%%%%%%%%%%%%%%%%%%%%%%%%%%%%%%
\section{PWA functions and Neural Networks}
    \begin{figure*}[t!]
    \centering
    \captionsetup{width=0.95\linewidth}
    \includegraphics[width=\linewidth]{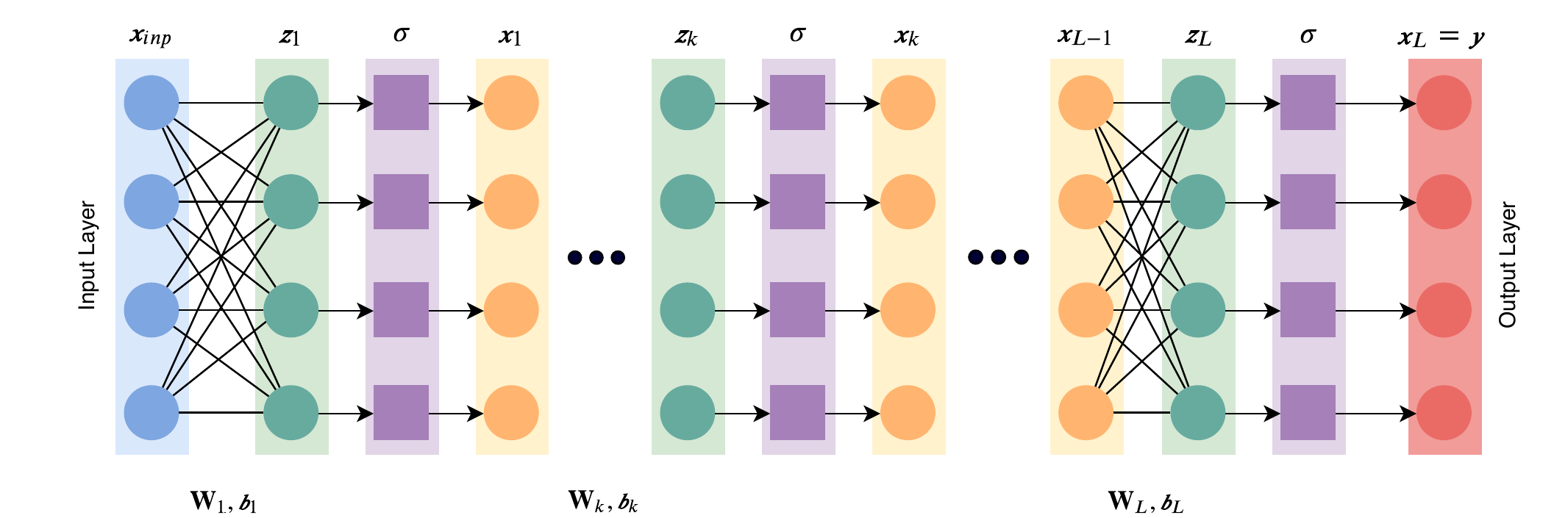}
    \caption{Neural network with $L$ fully connected layers with $\sigma$ as an activation function. The $k^{th}$ fully connected layer takes in the output of the previous layer $\bm{x}_{k-1}$ and produces $\bm{z}_k = \bmr{W}_k \bm{x}_{k-1} + \bm{b}_k$, where $\bm{x}_0 = \bm{x}_{inp}$. This value is then passed through the nonlinear activation function $\sigma$ which operates element-wise on $z_k$, yielding $\bm{x}_k = \sigma(\bm{z}_k)$.  }
    \label{fig:DNN}
    \end{figure*}
    
    Neural networks consisting of linear / affine layers and piecewise affine (PWA), continuous activation functions are themselves PWA and continuous \cite{DBLP:journals/corr/EldanS15} (see Equation \eqref{eq:pwa}). Note that any PWA function can also be written as a piecewise linear (PWL) function:
    
    \begin{equation}
    \label{eq:pwl}
    f(\bm{x}) = \bmat{\bm{w}^T_i & \bm{b}_i}\bmat{\bm{x} \\ 1} \;\;\;\; \forall \bm{x} \in \Omega_i
    \end{equation}
    
    This can also be viewed as expressing $\bm{x}$ in homogeneous coordinates. This allows chains of affine transformations to be written more compactly as a series of matrix multiplications.
    
    % For brevity, any vector $\bm{x}$ in homogeneous coordinates is written $\bar{\bm{x}}$.
    A general neural network can be viewed as a graph of nodes (neurons) with weighted, directed edges, where some of the nodes are taken to be inputs, and some are outputs. Each neuron is associated with a scalar activation function $\sigma$, and a scalar bias term $b$. The value of any given neuron is $\sigma(z)$, where $z$ is the sum of its weighted inputs, plus the bias term. The value of the neuron is passed on to other neurons through the outgoing edges. The values of the input neurons are set directly. In summary, for some neuron $i$, its value $x_i$ is:
    
    \begin{equation}
    \label{eq:single-layer-op}
        x_i = \sigma(z_i) = \sigma(\sum_{j \in \mathcal{P}(i)} (w_{ij} x_j + b_i)) 
    \end{equation}
    
    Where $\mathcal{P}(i)$ is the set of neurons that are connected to neuron $i$, and $w_{ij}$ is the weight of the edge from $j$ to $i$. This general formulation presupposes no structure, and may include cycles. A more common architecture is to assume that the neurons are organised into layers that are connected adjacently, but not internally, as can be seen in Fig. \ref{fig:DNN}. This is known as a \textit{feedforward neural network}. The advantage of this formulation is that the neurons of each layer can be grouped together into vectors $\bm{x}_k$, and the propagation of the input through the network can be expressed as matrix operations, where the connection weights $w_{ij}$ from Equation \eqref{eq:single-layer-op} are organised into a matrix $\bmr{W}_k$.
    
    \begin{equation}
    \label{eq:fully-connected}
    \bm{x}_k = \sigma(\bm{z}_k) = \sigma(\bmr{W}_k\bm{x}_{k-1} + \bm{b}_k)
    \end{equation}

    Such a layer is called \textit{fully-connected}, as every neuron in layer $k$ is connected to every neuron in layer $k-1$. The size of the matrix $\bmr{W}_k$ can become intractable for large $x_{k-1}$, for example when processing images. This issue can be mitigated by imposing some structure on the weights. For example, a \textit{convolutional layer} assumes that different parts of the input will be processed similarly ($\bmr{W}_k$ is redundant) such that Equation \eqref{eq:fully-connected} can be replaced by a convolution operation. These types of layers have proven to be instrumental for networks that operate on images.
    
    The activation function $\sigma$ is chosen to be any nonlinear function that maps $\mathbb{R}$ to some interval, and is applied element-wise to the layer input $\bm{x}_k$. Historically $\sigma$ has been selected as the sigmoid function $(1 + \exp(-x))^{-1}$, as this resembles the action potential exhibited in biological neurons. However, the sigmoid function is associated with the \textit{vanishing gradient problem} in deep networks, making it difficult to train \cite{bengio1994learning, pascanu2013difficulty}. A popular activation function that mitigates this issue is the Rectified Linear Unit (ReLU), a PWA function.
    
    % A commonly used PWA activation function is the Rectified Linear Unit (ReLU).
    \begin{equation}
    \label{eq:relu}
    \sigma(z) = \Bigg\{ \;\mat{z, \;\;\;\;\; z\geq 0 \\ 0, \;\;\;\;\; z<0}
    \end{equation}
    
    Most neural network implementations generalise these networks as computational graphs that operate on higher-dimensional tensors\footnote{Colour images can be represented with three dimensions, video with four (or five if sound is included).}, which greatly benefits the efficiency of evaluation and training. The different types of layers are thus generalised as operations on tensors, which may potentially be nonlinear. As previously mentioned, the scope of this work is limited to linear layers such as fully connected and convolutional layers, and to PWA activation functions. 
    
    % Before showing this, some notation is introduced. 
    %It is beneficial to think of a neural network as a \textbf{computational graph}, (\textbf{figure needed}), where each node accepts a vector of inputs, performs some operation, and outputs another vector. In this model, the connections between two layers of neurons can be seen as a node, and the activation functions that are applied at the end of each neuron can be seen as a node. 
    Consider a network with $L$ fully connected layers, as in Fig. \ref{fig:DNN}. The operation of each fully connected layer is an affine transformation, as shown in Equation \eqref{eq:fully-connected}. This can be converted to a linear transformation using Equation \ref{eq:pwl}:
    
    \begin{equation}
    \label{eq:layer-transformation}
    \bmat{\bm{z}_k \\ 1} = \bmat{\bmr{W}_k & \bm{b}_k \\ \bm{0} & 1} \bmat{\bm{x}_{k-1} \\ 1} = \bmr{T}_k \bmat{\bm{x}_{k-1} \\ 1}
    \end{equation}
    
    The last row of $\bmr{T}_k$ has been added to keep $\bm{z}_k$ in homogeneous coordinates. Relaxing the notation a little, the composition operator $\circ$ is allowed to operate on matrix multiplications such that: $(\bmr{T}_2 \circ \bmr{T}_1)\bm{x} = \bmr{T}_2\bmr{T}_1\bm{x}$. A feedforward neural network with input $\bm{x}$, output $\bm{y}$, and no branches can then be expressed as a series of alternating matrix multiplications and applications of $\sigma$.
            
    \begin{equation}
    \label{eq:compact-network}
    \bmat{\bm{y} \\ 1} = \Big(\sigma_n \circ \bmr{T}_n \circ \dots \circ \sigma_2 \circ \bmr{T}_2 \circ \sigma_1 \circ \bmr{T}_1\Big)\Bigg(\bmat{\bm{x} \\ 1}\Bigg)
    \end{equation}
    
    It is now easy to see that without nonlinear activation functions, this network would simplify into a single matrix multiplication. The nonlinearity of $\sigma$ is thus crucial to the representational power of neural networks. Equation \eqref{eq:compact-network} yields useful insights when attempting to obtain the PWA form of a neural network. To motivate this, consider a network with $L$ layers, each consisting of a single neuron with the following PWA activation function:
    
    \begin{equation}
    \label{eq:pwa-activation-example}
    \sigma(z) = \begin{cases}
        a_1 z, \;\;\;\;\; z \geq 0 \\
        a_2 z, \;\;\;\;\; z < 0
    \end{cases}
    \end{equation}
    
 %   \begin{figure}[t!]
 %   \centering
 %   \captionsetup{width=0.95\linewidth}
 %   \includegraphics[width=\linewidth]{fig/single-node-layers.png}
 %   \caption{Neural network with $L$ layers, each with one neuron. Here the activation functions and fully-connected layers have been drawn as separate nodes. Note that every edge between nodes has a weight $w_k$ that is not shown here. The operation performed by each $\Sigma$ node is then $w_k x_k + b_k$}
  %  \label{fig:single-node-layers}
   % \end{figure}
    
    This activation function has two linear regions, separated by $z=0$. The output of each fully connected layer is $z_k$, and the resulting activation $\sigma(z_k)$ is called $x_k$. The output $y$ of the network can be written:
    
    \begin{equation}
    \label{eq:pwa-activation}
    y = x_L = \begin{cases}
        a_1 z_{L}, \;\;\;\;\; z_{L} \geq 0 \\
        a_2 z_{L}, \;\;\;\;\; z_{L} < 0
    \end{cases}
    \end{equation}
    
    The activation function $\sigma$ thus splits its input into 2 separate regions. This expression can be expanded recursively, showing that the previous activation also splits its input in two, doubling the number of cases.
    
    \begin{equation}
    {\small
    y = x_{L} = \begin{cases}
        \,a_1^2 w_{L} z_{L-1} + a_1 b_{L}, \\
        \,\text{\textbf{for}} \; (z_{L} \geq 0) \land (z_{L-1} \geq 0) 
        \\
        \\
        \,a_1 a_2 w_{L} z_{L-1} + a_1 b_{L},\\
        \,\text{\textbf{for}} \; (z_{L}\geq0) \land (z_{L-1} < 0) 
        \\
        \\
        \,a_1 a_2 w_{L} z_{L-1} + a_1 b_{L},\\
        \,\text{\textbf{for}} \; (z_{L}<0) \land (z_{L-1} \geq 0) 
        \\
        \\
        \,a_2^2 w_{L} z_{L-1} + a_1 b_{L},\\
        \,\text{\textbf{for}} \; (z_{L}<0) \land (z_{L-1} < 0) 
    \end{cases}
    }
    \end{equation}
    
    Continuing the expansion leads to $2^L$ different cases, each one corresponding to the set of signs of all $z_k$, namely $(\sign(z_1), \sign(z_2), \dots, \sign(z_L))$. More generally, $z_k$ must lie in one of the intervals that the activation function $\sigma(z)$ is defined on, thus determining what $\sigma(z)$ is. This active interval is defined as the \textbf{activation} of a neuron. The set of activations of all neurons in a network is called the \textbf{activation pattern} \cite{serra2018bounding}, denoted by $\pi$. In the previous example, if the intervals of Equation \eqref{eq:pwa-activation} are defined as $(-, +) = (\{z\;|\;z<0\}, \{z\;|\;z\geq0\})$, then an activation pattern could have the form $\pi = (-, - ,+, \dots, -)$.
    
    The activation patterns are a natural way to characterise the linear regions of a neural network. Given some $\pi_i$, the corresponding case for each neuron can be selected (see Equation \eqref{eq:pwa-activation-example}), yielding the local PWA representation of the network. In terms of the previously introduced notation, this can be seen as substituting all $\sigma_i$ in \eqref{eq:compact-network} with their corresponding linear transformations, allowing the whole chain to simplify into a single matrix multiplication:
    
    \begin{gather}
    \begin{aligned}
    \label{eq:applying-pi}
    \Big(\sigma_n \circ \bmr{T}_n \circ \dots \circ \sigma_1 \circ \bmr{T}_1\Big)(\bmat{\bm{x} \\ 1}) \Bigg\vert_{\pi_i} = \bmr{P}_i\bmat{\bm{x} \\ 1}
    \end{aligned}
    \end{gather}
    
    However, not all activation patterns will be attainable by a given neural network. This can be seen in the next section in Fig. \ref{fig:simple-network}. The challenge is then to identify all the valid activation patterns $\pi_i$, find the corresponding affine transformation $\bmr{P}_i$, and assign to them the corresponding region $\mathcal{R}_i$ of the input space. The situation is also complicated further when the layers are allowed to contain more than one neuron. The next section takes an iterative approach to this problem, by considering some simple examples that build on each other. 
    
    % The scope of the work is limited to networks using the popular ReLU activation function.
    To this end, the main contribution of this work is an algorithm that can convert any DNN using fully connected layers and ReLU activations into its exact PWA representation that can be visualised and analysed, giving an insight into the inner workings of the network. This was achieved by utilising existing linear programming (LP) methods (specifically the MPT toolbox for MATLAB\supercopyright, see\cite{MPT3}) for working with polyhedral sets and hyperplane arrangements. The approach can also be extended to any linear / affine layer (convolutional layers, batch normalisation), as well as any PWA activation function (Leaky ReLU, maxout).

%%%%%%%%%%%%%%%%%%%%%%%%%%%%%%%%%%%%%%%%%%%%%%%%%%%%%%%%%%%%%%%%%%%
\section{PWA Representation of a Simple Neural Network with ReLU Activations}
    
\begin{figure}[t!]
        \centering
        \captionsetup{width=0.95\linewidth}
        
        \begin{subfigure}{\linewidth}
            \includegraphics[width=\linewidth]{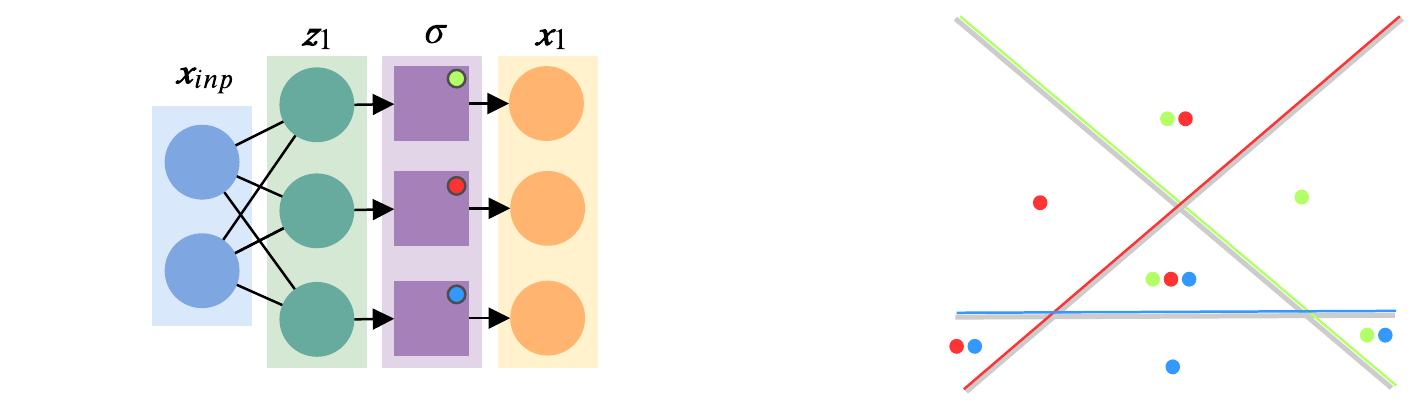}
            \caption{Single hidden layer}
            \label{fig:simple_network_1_layer}
        \end{subfigure}
        
        \begin{subfigure}{\linewidth}
            \includegraphics[width=\linewidth]{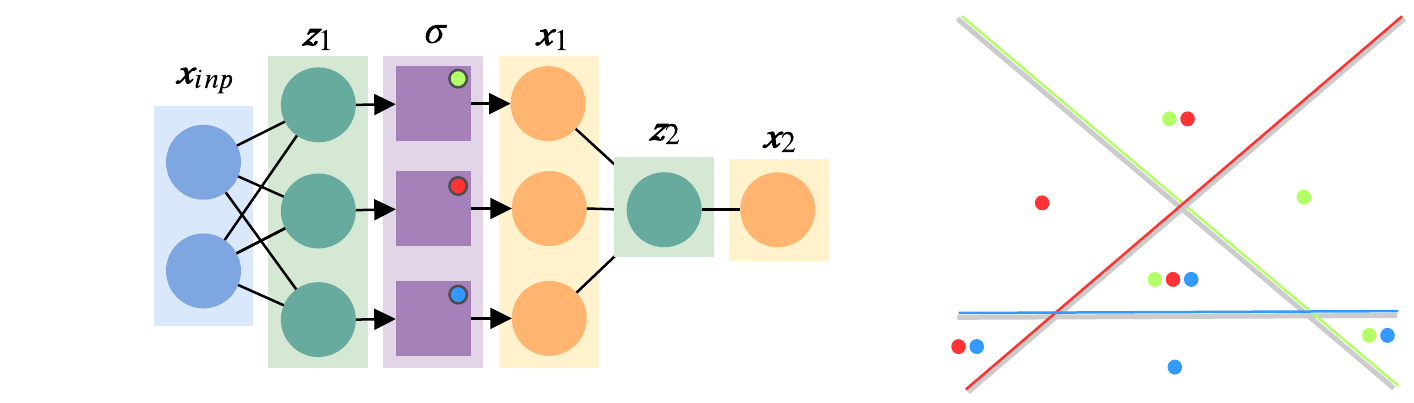}
            \caption{Two hidden layers, no activation on the second}
            \label{fig:simple_network_1_layer_1_output}
        \end{subfigure} 
        
        \begin{subfigure}{\linewidth}
            \includegraphics[width=\linewidth]{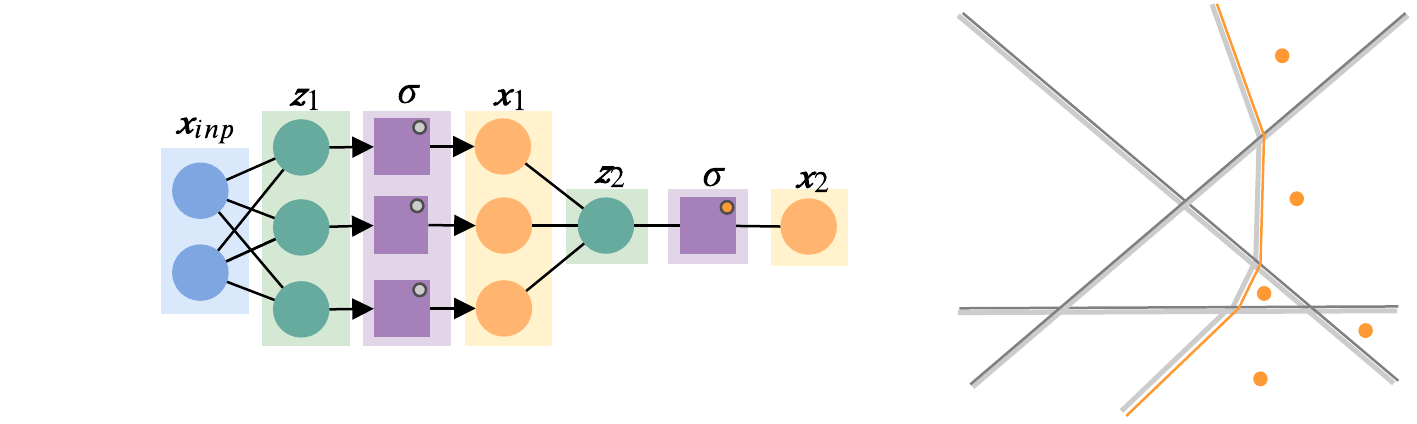}
            \caption{Two hidden layers with activation}
            \label{fig:simple_network_2_layer}
        \end{subfigure} 
        
        \caption{
        The linear regions of three successively larger networks. Each neuron in the first hidden layer has boundary in the form of a line. The activation patterns corresponding to each region have been given as coloured dots, where the absence of a dot implies that a neuron is inactive. Adding a fully-connected layer with no activations does not affect the linear regions, as shown in (b). Adding an activation function to a fully-connected layer adds additional boundaries for each neuron, as shown in (c). Here there is only one neuron in the last layer. The boundary is different in each linear region of the first layer, but remains continuous. The boundary thus appears to bend when intersecting with the boundaries of previous layers. }
        
        \label{fig:simple-network}
        
    \end{figure}

    Consider a neural network with 2 inputs and 1 hidden layer with 3 nodes and ReLU activation, as shown in Fig. \ref{fig:simple_network_1_layer}. The general form of the network is:
            
    \begin{equation}
    \label{eq:simple_network_1_layer}
        \bm{f}(\bm{x}) = \sigma(\bmr{T}\bmat{\bm{x} \\ 1}) = \sigma\bigg(\bmat{\bmr{W} & \bm{b} \\ 0 & 1}\bmat{\bm{x} \\ 1}\Bigg)
    \end{equation}
    
    The activation function $\sigma$ is ReLU, as given in Equation \eqref{eq:relu}. Equation \eqref{eq:simple_network_1_layer} can then be written as:
    
    \begin{equation}
    \label{eq:simple_network_1_layer_expanded}
        \bm{f}(\bm{x}) =\bmat{
        \max(0,\bm{w}^T_1\bm{x}+\bm{b}_1)) \\
        \max(0,\bm{w}^T_2\bm{x}+\bm{b}_2)) \\
        \max(0,\bm{w}^T_3\bm{x}+\bm{b}_3)) \\
        1
        }
    \end{equation}
    
\begin{figure}[t!]
        \centering
        \includegraphics[width=\linewidth]{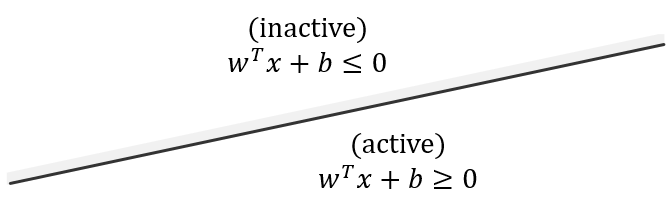}
        \captionsetup{width=0.95\linewidth}
        \caption{Each node with ReLU activation has two modes: one where it is active and one where it is inactive. Boundaries are therefore drawn with a shaded side representing the inactive side. }
        \label{fig:hyperplane-shading}
    \end{figure}

    The vector $\bm{w}^T_i$ represents the $i^{\text{th}}$ row of $\bmr{W}$. Each element of $\bm{f}(\bm{x})$ corresponds to the output of a neuron. Each neuron thus has two modes: one where the output is clipped to zero (because $\bm{w}^T\bm{x}+\bm{b} \leq 0$), and one where the output is $\bm{w}^T\bm{x}+\bm{b}$. The boundary between these two modes is given by $\bm{w}^T\bm{x}+\bm{b} = 0$, which defines a line. More generally, this boundary will be a hyperplane when there are more than two inputs. Each neuron thus bisects the input space, only outputting a positive, nonzero value if the input point $\bm{x}$ is on the positive side of the corresponding boundary. To visualise this directionality, the boundaries are drawn with a shaded side, as can be seen in Fig. \ref{fig:hyperplane-shading}.
            
    % Applying the ReLU function results in the negative elements of the output being clipped to zero. The effect of the ReLU nonlinearity is therefore better described as a \textit{deactivation function}, although this term will not be used to avoid confusion. 
    The arrangement of these boundaries defines a set of polyhedral regions $\mathcal{R}_i$ in the input space, each corresponding to a different activation pattern $\pi_i$. From \eqref{eq:simple_network_1_layer} it can be seen that an inactive neuron is equivalent to setting the corresponding row in the $\bmr{T}$ matrix to zero. The function in each region is therefore described by its own copy of the $\bmr{T}$ matrix, but with inactive rows being set to zero, as shown explicitly in Table \ref{tab:simple_network_1_layer}. Note that the activation function $\sigma(\cdot)$ has been completely removed from the expression.
    
    As there are no further layers, Table \ref{tab:simple_network_1_layer} is the complete PWA representation of the simple network in Fig. \ref{fig:simple_network_1_layer}, where the regions are defined implicitly. Section \ref{sec:get-regions} describes how to find and how to explicitly represent the regions. 

\begin{table}
    \centering
    \begin{tabular}{|c|c|}
        \hline
        Activation pattern $\pi$ & Computed PWA Function \\ \hline
        & \\
        \tikzdot[green, fill=green]{1.5pt}
        & ${\tiny \bmat{\bm{w}_1^T & b_1 \\ \bm{0} & 0 \\ \bm{0} & 0 \\ \bm{0} & 1}\bmat{\bm{x} \\ 1}}$  \\ 
        & \\ \hline
        & \\
        \tikzdot[red, fill=red]{1.5pt}
        &  ${\tiny\bmat{\bm{0} & 0 \\ \bm{w}_2^T & b_2 \\ \bm{0} & 0 \\ \bm{0} & 1}\bmat{\bm{x} \\ 1}}$   \\
        & \\ \hline
        & \\
        \tikzdot[blue, fill=blue] {1.5pt}
        &  ${\tiny\bmat{\bm{0} & 0 \\ \bm{0} & 0 \\ \bm{w}_3^T & b_3 \\ \bm{0} & 1}\bmat{\bm{x} \\ 1}}$  \\ 
        & \\ \hline
        & \\
        \tikzdot[green, fill=green]{1.5pt}\,\tikzdot[red, fill=red]{1.5pt}
        &  ${\tiny\bmat{\bm{w}_1^T & b_1 \\ \bm{w}_2^T & b_2 \\ \bm{0} & 0 \\ \bm{0} & 1}\bmat{\bm{x} \\ 1}}$  \\ 
        & \\ \hline
        & \\
        \tikzdot[green, fill=green]{1.5pt}\,\tikzdot[blue, fill=blue]{1.5pt} 
        &  ${\tiny\bmat{\bm{w}_1^T & b_1 \\ \bm{0} & 0 \\ \bm{w}_3^T & b_3 \\ \bm{0} & 1}\bmat{\bm{x} \\ 1}}$  \\ 
        & \\ \hline
        & \\
        \tikzdot[red, fill=red]{1.5pt}\,\tikzdot[blue, fill=blue]{1.5pt} 
        &  ${\tiny\bmat{\bm{0} & 0 \\ \bm{w}_2^T & b_2 \\ \bm{w}_3^T & b_3 \\ \bm{0} & 1}\bmat{\bm{x} \\ 1}}$  \\ 
        & \\ \hline
        & \\
        \tikzdot[green, fill=green]{1.5pt}\,\tikzdot[red, fill=red]{1.5pt}\,\tikzdot[blue, fill=blue]{1.5pt}
        &  ${\tiny\bmat{\bm{w}_1^T & b_1 \\ \bm{w}_2^T & b_2 \\ \bm{w}_3^T & b_3 \\ \bm{0} & 1}\bmat{\bm{x} \\ 1}}$  \\
        & \\\hline
    \end{tabular}
    \captionsetup{width=0.95\linewidth}
    \caption{The complete PWA representation for the network in figure \ref{fig:simple_network_1_layer}. Each region computes its own transformation, with some of the rows zeroed out.}
    \label{tab:simple_network_1_layer}
\end{table}

    % \label{sec:algo:examples:output}    
    % \begin{figure}[H]
    %     \centering
    %     \includegraphics[width=0.25\linewidth]{fig/simple_network_1_layer_1_output.png}
    %     \captionsetup{width=0.7\linewidth}
    %     \caption{Adding another layer with no activation will not effect the linear regions at all. However, it will modify the $\bm{P}$ matrix of each region defined by the previous layers.}
    %     \label{fig:simple_network_1_layer_1_output}
    % \end{figure}
    
    Adding an output layer with no activation is straightforward. The resulting network is shown in Fig. \ref{fig:simple_network_1_layer_1_output}. The second layer has no activation, and computes the function:
    
    \begin{gather}
    \begin{aligned}
    \label{eq:simple_network_1_layer_1_output}
        \bmat{\bm{f}(\bm{x}) \\ 1} &= \bmr{T}_2\bmat{\bm{x} \\ 1} = \bmat{\bmr{W}_2 & \bm{b}_2 \\ 0 & 1}\bmr{T}_1\bmat{\bm{x} \\ 1} \\
        &= \bmat{\bmr{W}_2\bmr{W}_1 & \bmr{W}_2 \bm{b}_1 + \bm{b}_2 \\ \bm{0} & 1} \bmat{\bm{x} \\ 1}
    \end{aligned}
    \end{gather}
    
    where $\bmr{T}_i$ represents the transformation matrix corresponding to region $\mathcal{R}_i$, as shown in table \ref{tab:simple_network_1_layer}. The effect of adding another layer with no activation is just a matrix multiplication between the $\bmr{T}_k$ matrix of the new layer and the $\bmr{P}_i$ matrices of each region $\mathcal{R}_i$. This shows that adding layers without activation functions will not affect the linear regions.
    
    % \subsubsection{Example 3: Adding ReLU activation}
    % \label{sec:algo:examples:morelayers}    
    
    % \begin{figure}[H]
    %     \centering
    %     \includegraphics[width=0.8\linewidth]{fig/simple_network_2_layer.png}
    %     \captionsetup{width=0.7\linewidth}
    %     \caption{Simple network with 2 hidden layers. Note how the last node bisects each of its active regions in a different way, yet remains continuous across regions, such that it appears to bend at the boundaries of the previous layer.}
    %     \label{fig:simple_network_2_layer}
    % \end{figure}
    An activation function is now added to the last neuron of the network in Fig. \ref{fig:simple_network_1_layer_1_output}. The ReLU function deactivates the node in certain regions, switching it on and off. As before, there is a boundary that describes this switching behaviour. However, this time the input space consists of multiple regions defined by the previous layers. Importantly, the parameter matrix $\bmr{T}_i$ for each region is different, implying that the boundary introduced by the last node will be different for each region. The result will be similar to what is depicted in Fig. \ref{fig:simple_network_2_layer}. 
    
\begin{figure}[t!]
        \centering
        \includegraphics[width=\linewidth]{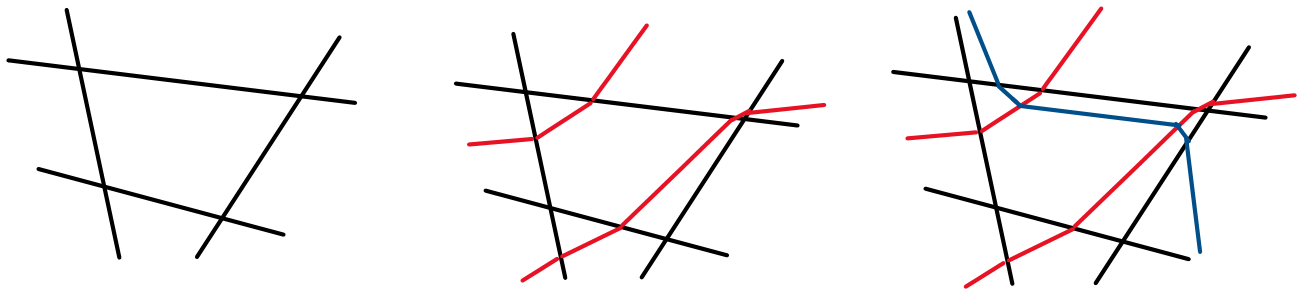}
        \captionsetup{width=0.95\linewidth}
        \caption{Consider the activations boundaries of this new neural network. The boundaries of each neuron "bend" at the boundaries of previous layers.}
        \label{fig:algo:bending_boundaries}
    \end{figure}

    The new boundary is continuous across the boundaries of the previous layers, but it will "bend" as it crosses them. This pattern continues as more layers are added, as new boundaries will bend when intersecting the boundaries of all previous layers. Another example of this structure is given in Fig. \ref{fig:algo:bending_boundaries}.
    
    Applying the ReLU nonlinearity is relatively simple. Every region found so far is associated with its own unique $\bmr{T}$ matrix, which defines the output of each neuron in that region. The procedure from the first example can then be applied separately to each region, yielding the new set of regions. The subregions inherit their parents $\bmr{T}$ matrix, with the inactive rows zeroed out.
    
    With this in place, the process of converting the network to its PWA form can be generalised to any number of layers. The missing piece is a method to find the regions defined by each layer.

\section{Finding the Linear Regions of a Neural Network}
\label{sec:get-regions}
    The previous examples demonstrated how the PWA representation may be obtained when the activation pattern $\pi_i$ is known, and described the structure of the linear regions. What now remains is to explicitly compute the regions, which are polyhedra. It is most convenient to define the regions using the hyperplanes themselves. This is known as the H-representation, where the region is defined as the intersection of the half-spaces defined by the hyperplanes \cite{fukuda2004faq}. If the bounding hyperplanes have indices $\mathcal{H} = \{1, 2, \dots, 3 \}$, then the polyhedron $\mathcal{P}$ can be written as:
    
    \begin{equation}
    \label{eq:H-repr}
        \mathcal{P} = \{\,\bm{x}  \;\vert\; \bm{w}_i^T\bm{x} + b_i \geq 0, \;\forall i \in \mathcal{H}\,\}
    \end{equation}
    
    Alternatively, this can be written as the matrix inequality:
    
    \begin{equation}
    \label{eq:H-repr-matrix}
        \bmr{H} \bmat{\bm{x} \\ 1} = \bmat{\bm{w}_1^T & b_1 \\ \vdots & \vdots \\ \bm{w}_n^T & b_n}\bmat{\bm{x} \\ 1} \geq 0
    \end{equation}
    
    In particular, the $\bmr{H}$ matrix representation allows one to easily test whether a point is contained within the region, to compute an internal point by finding the Chebyshev centre, and can be used to check for intersections between polytopes of different dimensions \cite{baotic2009polytopic, MPT3}. A drawback is that there may be redundant constraints, which can slow down later operations. It is also generally expensive to identify and remove the redundant constraints, as this involves solving an LP for each hyperplane, although there exist heuristics to reduce this number \cite{marechal2017efficient}.
    
    The issue of finding the regions defined by the neuron boundaries of a layer with PWA activation is equivalent to finding the regions of a hyperplane arrangement. A compact and rigorous discussion of hyperplane arrangements is given in \cite{introToHplaneArrangements}. An important result is an upper bound on the maximal number of regions for $n$ hyperplanes in $\mathbb{R}^d$ \cite{zavslavsky-facing-up}.
        
    \begin{equation}
    \label{eq:zavslavsky}
        \sum^d_{j=0} \binom{n}{j}
    \end{equation}
        
\begin{figure}[t!]
        \centering
        \captionsetup{width=0.95\linewidth}
        \includegraphics[width=\linewidth]{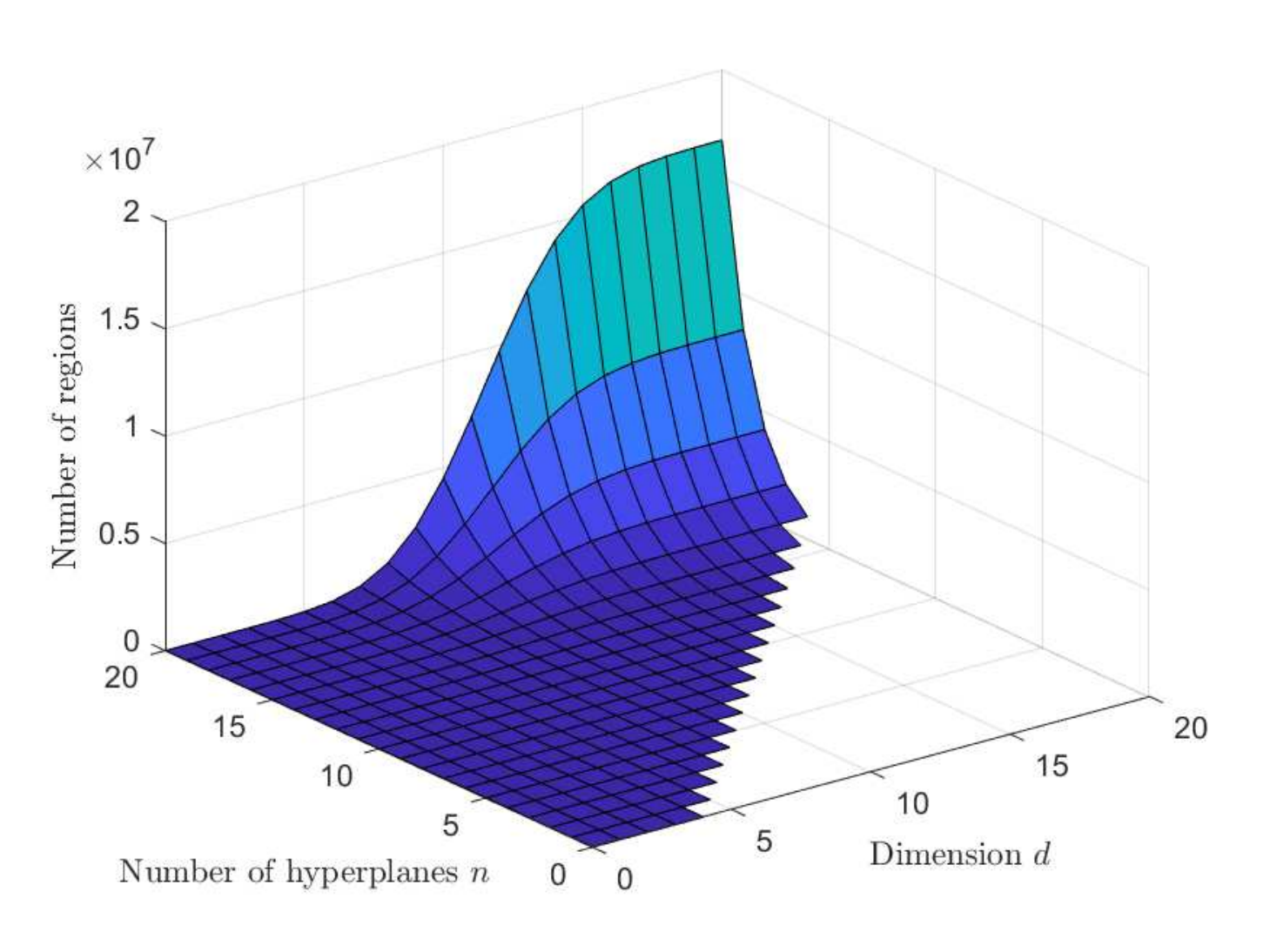}
        \caption{Growth of Zavslavsky's upper bound for the number of regions in a hyperplane arrangement.}
        \label{fig:zavslavsky_growth}
    \end{figure}

    This expression grows quickly with both $d$ and $n$, but not exponentially. A surface plot of \eqref{eq:zavslavsky} is given in Fig. \ref{fig:zavslavsky_growth}. In practice the number of regions will be significantly less than this.
    
\begin{figure}[t!]
        \centering
        \captionsetup{width=0.95\linewidth}
        \includegraphics[width=0.8\linewidth]{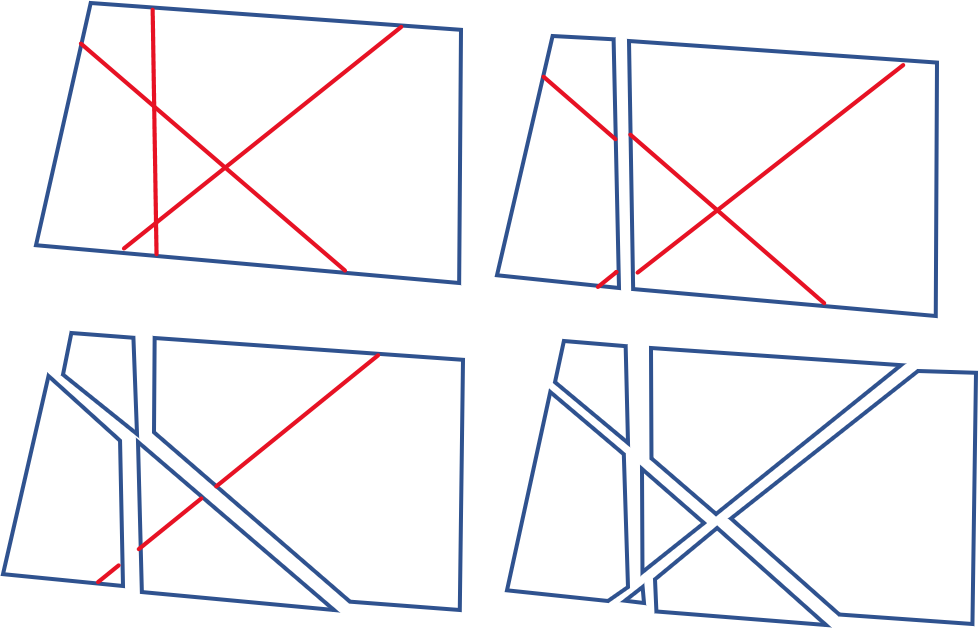}
        \caption{Illustration of a procedure for finding the regions of a hyperplane arrangement. Each hyperplane is considered in turn, and is used to bisect the previously found regions by adding it to their H-representations. It is necessary to search the previously found regions for intersections. The search space can be significantly reduced by checking the parent regions first. This can be accomplished by storing the regions in a binary tree structure, adding new nodes every time a region is bisected by a hyperplane.}
        \label{fig:repartition}
    \end{figure}

    The regions may be found by iteratively bisecting a growing collection of regions, as illustrated in Fig. \ref{fig:repartition}. This is done by adding the bisecting hyperplane as a constraint to the H-representation of the ``parent" region. If the bisecting hyperplane is given by $\bm{w}_{bi}^T\bm{x} +\bm{b}_{bi} = 0$, and the region is represented by $\bmr{H} = \bmat{\bmr{W} & \bm{b}}$, then the new $\bmr{H}$ matrices of the two subregions will be:
    
    \begin{gather}
    \begin{aligned}
    \label{eq;subregions}
        \bmr{H}^+ &= \bmat{\bmr{W} & \bm{b}\\ \bm{w}_{bi}^T & b_{bi}} \\
        \bmr{H}^- &= \bmat{\bmr{W} & \bm{b} \\ -\bm{w}_{bi}^T & -b_{bi}} 
    \end{aligned}
    \end{gather}
    
    If the hyperplane does not intersect the parent region, then the previous matrices describe empty sets. The intersection may be checked with a quick feasibility LP using Equation \eqref{eq:H-repr-matrix} as a constraint. The hyperplanes can then be considered one by one, checking for intersections with all of the regions found so far, and bisecting when there is an intersection.
    
    The main source of complexity associated with this procedure is the increasing number of regions that must be checked for intersections. The search space can be reduced significantly by retaining the parent regions found after each iteration and checking these instead. If a hyperplane does not intersect a region, then it will not intersect any of its subregions either. This procedure is shown in Alg. \ref{algo:repartition}.

\begin{algorithm}
\scriptsize
\caption{\small Procedure to obtain regions of hyperplane arrangement}
\SetAlgoLined
\DontPrintSemicolon

\SetKw{KwRow}{row}
\SetKwProg{Fn}{fn}{ is}{end}
\SetKwFunction{FMain}{main}
\SetKwFunction{FRepartition}{get\_regions}
\SetKwFunction{FSearch}{search}
\SetKwFunction{FLeft}{left\_child}
\SetKwFunction{FRight}{right\_child}
\SetKwFunction{FLayerCompose}{layer\_compose}
\SetKwFunction{FCalculateP}{calculate\_P}
\SetKwFunction{FIntPoint}{interior\_point}
\SetKwFunction{FZip}{zip}
 
\Fn{\FRepartition{initial region $\mathcal{R}_0$, hyperplanes $\mathcal{H}$}}{
    \Fn{\FSearch{region $\mathcal{R}$, hyperplane $h$}}{
        \tcc{Extract $\bm{w}$ and $b$ from $h$}
        $\bm{w}, b \leftarrow h = \big\{\,\bm{x} \;\;|\;\; \bm{w}^T\bm{x} = -b \, \big\}$\;
        \uIf{$\mathcal{R} \cap h \neq \emptyset$}{
            \eIf{$\mathcal{R}$ has no children}{
                \FLeft{$\mathcal{R}$} $\leftarrow \mathcal{R} \,\cap\, \big\{\,\bm{x} \;\;|\;\; \bm{w}^T\bm{x} > -b \, \big\}$\;
                \FRight{$\mathcal{R}$} $\leftarrow \mathcal{R} \,\cap\, \big\{\,\bm{x} \;\;|\;\; \bm{w}^T\bm{x} < -b \, \big\}$\;
            }{
                \FSearch{\FLeft{$\mathcal{R}$}, $h$}\;
                \FSearch{\FRight{$\mathcal{R}$}, $h$}\;
            }
        }
    }
    \ForEach{$h \in \mathcal{H}$}{
        \FSearch{$\mathcal{R}_0$, $h$}\;
    }
    \KwRet $\mathcal{R}_0$\;
}
\label{algo:repartition}
\end{algorithm}

\section{Algorithm}

    As shown in the examples, a neural network can be converted to its PWA representation in an iterative fashion, starting at the input layer. Each subsequent layer is then applied to each of the previously found regions. If the layer is a linear transformation (fully connected layer), the PWA transformation in each region is modified. If the layer is a ReLU activation function, each region is further subdivided by finding the node boundaries and solving the hyperplane arrangement problem.
    
    % The algorithm can be summarised as:
    
    % \begin{enumerate}
    %     \item Get the next layer
    %     \item Multiply the $\bm{P}$ matrices of all currently known regions with the $\bm{P}$ matrix of the layer
    %     \item If the layer has an activation (ReLU), iterate over all known regions and solve the resulting hyperplane arrangement problem.
    %     \item The $\bm{P}$ matrices of all new regions are updated by finding interior points and checking which rows of $\bm{P}$ are inactive. The inactive rows are set to zero.
    %     \item If there are more layers, go to point 1. Otherwise, return the set of known regions.
    % \end{enumerate}

    The set of currently known regions and their corresponding $\bmr{P}_i$ matrices is called the \textbf{working set} $\mathcal{W}$. Every element in $\mathcal{W}$ will be a tuple of the form $(\mathcal{R}_i, \bmr{P}_i)$, where $\mathcal{R}_i$ is a polyhedral region and $\bmr{P}_i$ is a matrix that defines the affine transformation computed within that region.
    
    The neural network $\mathcal{N}$ itself is represented as a sequence of nodes, which can either be fully connected layers (represented as the linear transformation $\bmr{T}$) or ReLU activations. The algorithm is presented in Alg. \ref{algo:pwa-overview}.
    
     \begin{algorithm}
\scriptsize
\caption{Convert neural network $\mathcal{N}$ to its PWA representation}
\SetAlgoLined
\DontPrintSemicolon

\SetKw{KwRow}{row}
\SetKwProg{Fn}{fn}{ is}{end}
\SetKwFunction{FMain}{pwa}
\SetKwFunction{FRepartition}{get\_regions}
\SetKwFunction{FLayerCompose}{layer\_compose}
\SetKwFunction{FCalculateP}{calculate\_P}
\SetKwFunction{FIntPoint}{interior\_point}
\SetKwFunction{FZip}{zip}

\KwResult{$\mathcal{W} = \big\{(\textit{region } \mathcal{R}_1, \textit{transformation }\bmr{P}_1), \dots, (\mathcal{R}_n, \bmr{P}_n) \big\}$}

\Fn{\FMain{network $\mathcal{N}$}}{
    $d = \text{input dimension of } \mathcal{N}$
    \textit{working set} $\mathcal{W} \leftarrow (\mathbb{R}^d, \bmr{I}_{d+1})$\;
    \ForEach{layer $N \in \mathcal{N}$}{
        \uCase{$N = $ fully connected layer with transformation $\bm{T}$}{
                \For{$(\mathcal{R}_k, \; \bmr{P}_k) \in \mathcal{W}$}{
                    $\bmr{P}_k \leftarrow \bmr{T} \bmr{P}_k$
                }
            }
        \Case{$N = $ ReLU activation}{
            $\mathcal{W}_{new} \leftarrow \emptyset$\;
            \For{$(\textit{region }\mathcal{R}_k, \; \textit{transformation }\bmr{P}_k) \in \mathcal{W}$}{
                \textit{subregions} $\mathcal{S}_{new} \leftarrow$ \FRepartition{$\mathcal{R}_k$, $\bmr{P}_k$}\;
                \For{subregion $\mathcal{S}_j \in \mathcal{S}$}{
                    $\bm{v}_{int}$ = \FIntPoint{$\mathcal{S}_i$}\;
                    \textit{inactive rows} $\bm{r} = \{\, i \;\vert\; [\bmr{P}_k \bm{v}_{int}]_i < 0 \,\}$\;
                    \textit{subregion matrix} $\bmr{P}_{new} \leftarrow \bmr{P}_i$, with inactive rows $\bm{r}$ set to zero\;
                    $\mathcal{W}_{new} \leftarrow \mathcal{W}_{new} \;\Vert\; \{\, (\mathcal{S}_{j}, \mathcal{\bm{P}}_{new}) \,\}$\;
                }
            }
            $\mathcal{W} \leftarrow \mathcal{W}_{new}$
        }
    }
}
\label{algo:pwa-overview}
\end{algorithm}

    As the size of the working set will increase after processing each layer, it is clear that the worst case performance of the algorithm will depend greatly on the total depth of the network. However, it is not clear how quickly the working set will grow. For example, some regions in the working set may be intersected multiple times by the node boundaries in the next layer, while others will not be intersected at all. Despite this, the problem is inherently parallelisable. When parsing a layer of the network, the hyperplane arrangement problem is solved separately for each region in the working set, allowing for significant speedups when many cores are available. As will be shown in the next section, the number of regions in the working set quickly becomes very large, suggesting that the algorithm could benefit greatly from GPU hardware acceleration.
     
    % Collections (discrete sets) of parameter matrices $\bm{P}_i$ and regions $D_i$ are labeled $\mathcal{P}$ and $\mathcal{D}$ respectively.

\section{Results}
\label{sec:algo:runtime}

\begin{figure}[t!]
        \centering
        \begin{subfigure}{\linewidth}
            \includegraphics[width=\linewidth]{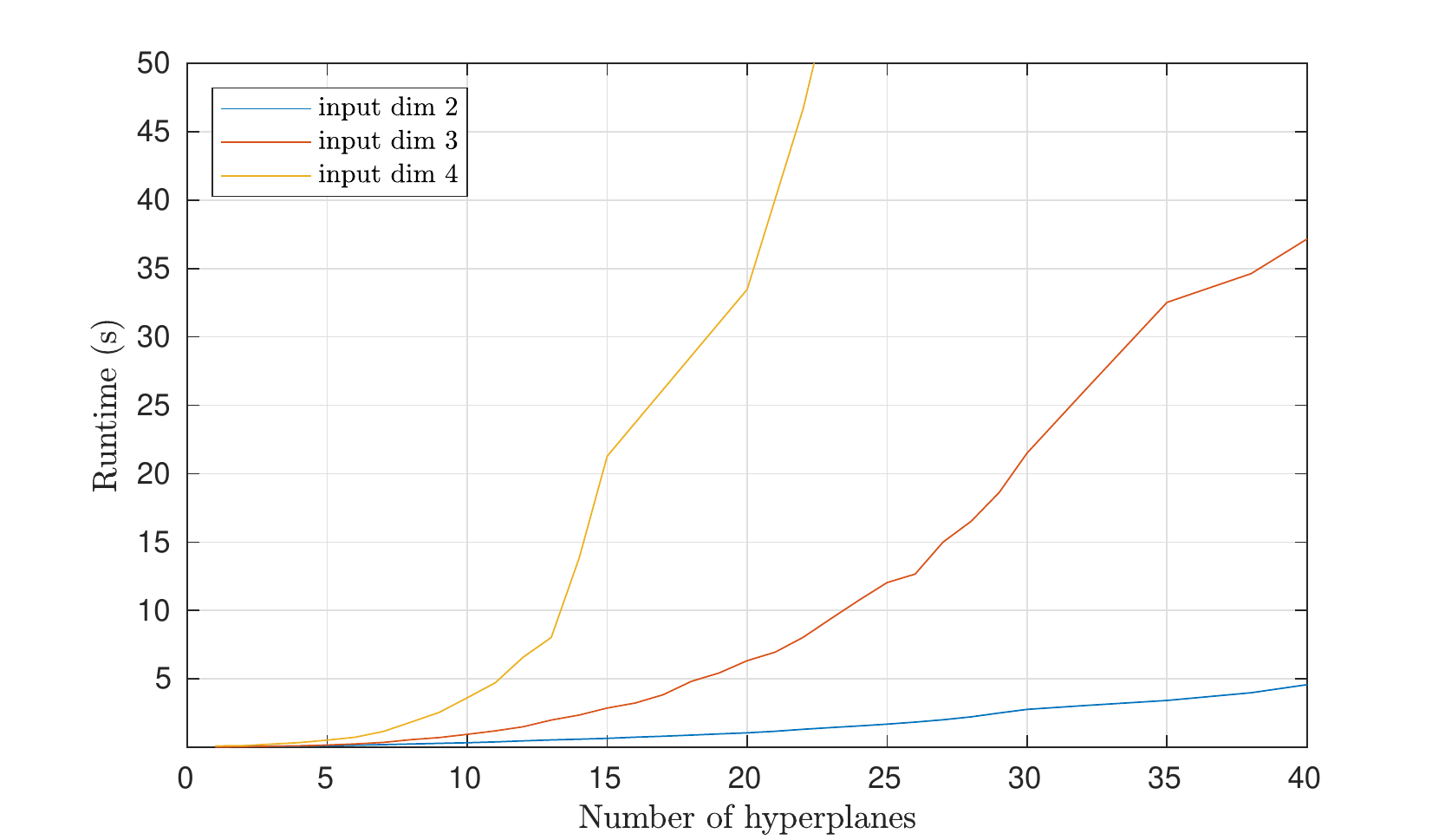}
            \caption{Runtime of \textbf{get\_regions()} against the number of hyperplanes}
            \label{fig:algo:repart_runtime_vs_hyperplanes}
        \end{subfigure} \hspace{5mm}
        \begin{subfigure}{\linewidth}
            \includegraphics[width=\linewidth]{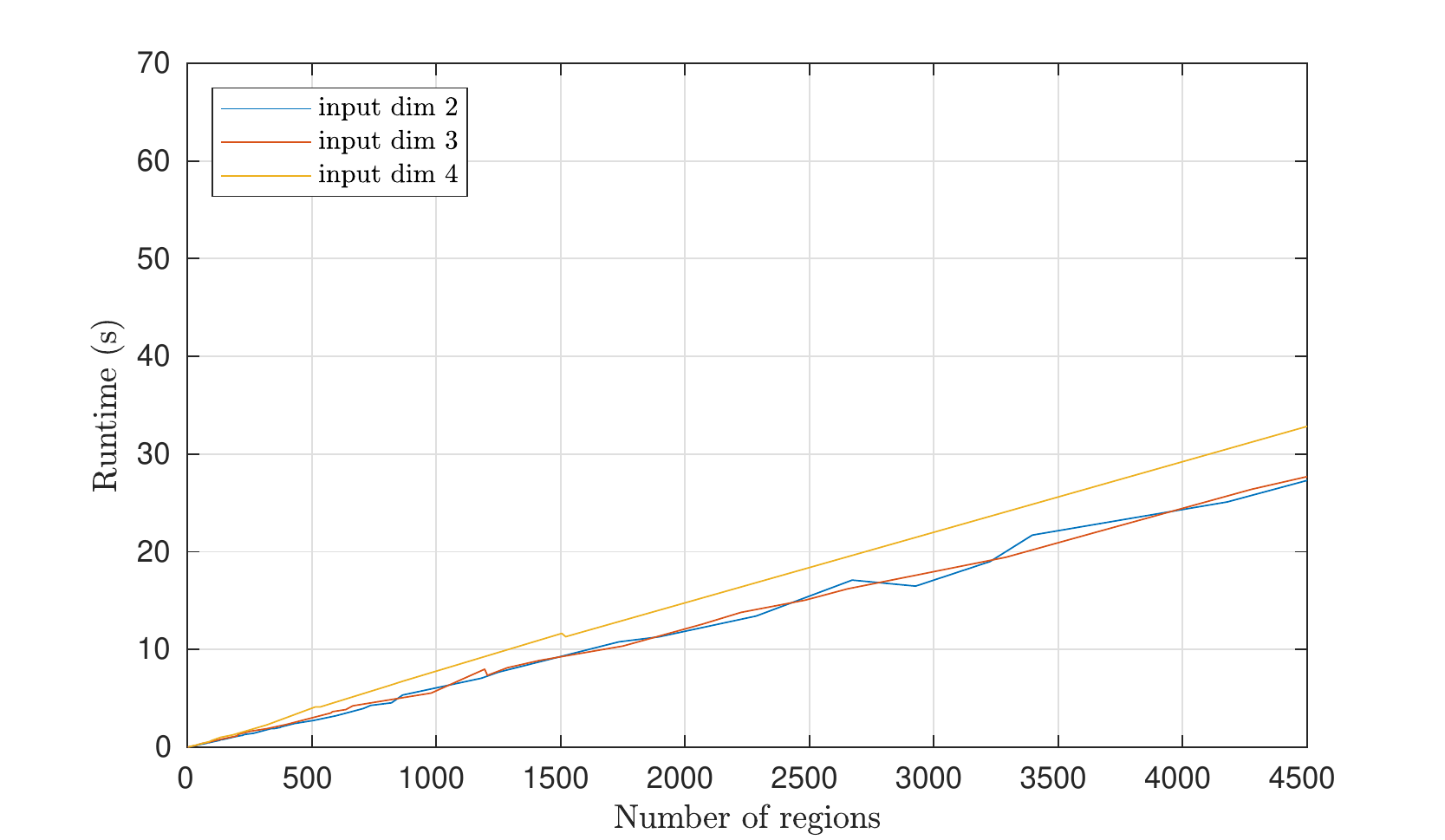}
            \caption{Runtime of \textbf{get\_regions()} against the number of regions}
            \label{fig:algo:repart_runtime_vs_regions}
        \end{subfigure}

        \caption{The runtime of \textbf{get\_regions()} increases significantly with the number of nodes/hyperplanes, and appears to be somewhat linear with respect to the number of regions found. Increasing the input dimension increases the number of regions significantly, and adds a small, per-region cost that scales with the size of the arrangement.}
        \label{fig:algo:repart_runtime}
    \end{figure}

    All runtimes were measured using a machine with a 6-core, 3,5 GHz processor and 16 GB of RAM. The polyhedral computations described in previous sections were performed using the MPT toolbox for MATLAB\supercopyright. The results for \textbf{get\_regions()} in terms of the number of hyperplanes have been presented together in Fig. \ref{fig:algo:repart_runtime_vs_hyperplanes}. The runtimes for \textbf{get\_regions()} are also presented in terms of the number of regions in Fig. \ref{fig:algo:repart_runtime_vs_regions}, showing that the runtime is roughly proportional to the number of regions found. More surprisingly, the effect of increasing the input dimension (and thus the size of the required LPs) is almost negligible in comparison. This suggests that it is the high number of calls to the LP solver, rather than the size of the LPs, that dominates the time complexity of \textbf{get\_regions()}. As the LPs are generally quite small, choosing an LP solver with a low amount of presolving might yield significant improvements. The author's implementation used the default LP solver included with MATLAB\supercopyright (\textit{linprog()}), which is often outperformed by other solvers. 
    
\begin{figure}[t!]
        \centering
        \captionsetup{width=.95\linewidth}
        \includegraphics[width=\linewidth]{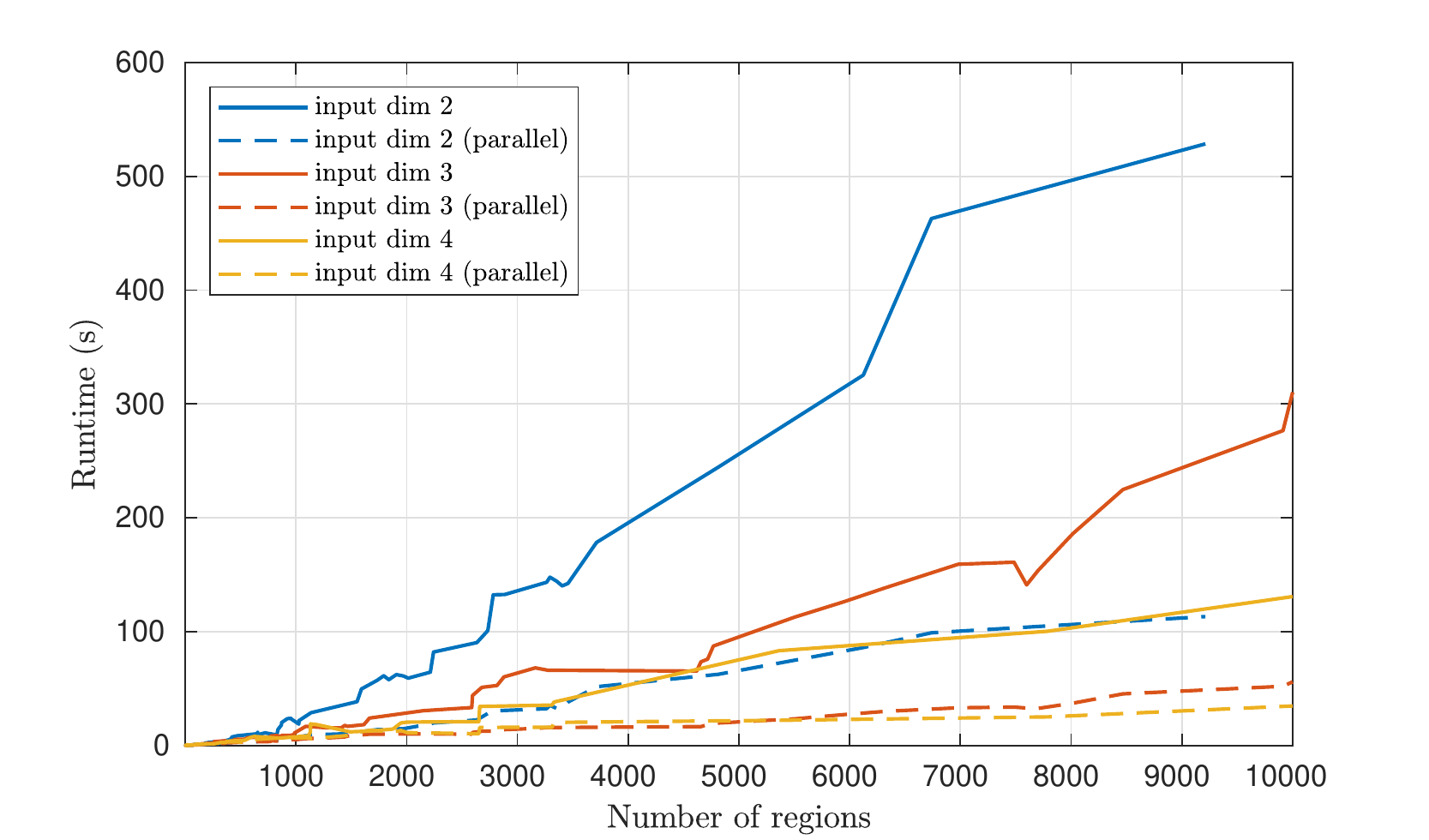}
        \caption{Runtime of the main algorithm against the number of regions found, with and without parallelisation. }
        \label{fig:convert_runtimes}
    \end{figure}

    The runtime of the main algorithm was measured with and without parallelisation on the available 6 cores. The runtime as a function of the number of regions of the final network is shown in Fig. \ref{fig:convert_runtimes}. Networks with an input dimension of up to four were processed, as the number of regions quickly exploded and the runtimes became intractable for networks with larger input dimensions. The runtime increases exponentially with the size of the network. Parallelisation was very effective, with the performance increasing by a factor approaching the number of cores used (6 cores). The per-region cost decreases with the input dimension, suggesting that there is an efficiency gain when increasing the input dimension. However, for any given number of regions, the corresponding points on the lines represent networks of very different sizes. For example, a network with two inputs and three hidden layers with width 10 might have a similar number of regions as a network with four inputs and two hidden layers with width 5. However, the first network will likely take longer to convert due to it having an additional hidden layer, and wider layers overall. 
    
    % \begin{figure}[t!]
    %     \centering
    %     \captionsetup{width=0.95\linewidth}
    %     \includegraphics[width=\linewidth]{fig/algorithm/repartition_intersection_checks_compact.eps}
    %     \caption{Number of intersection checks performed by \textbf{get\_regions()}. The curves appear to match the runtimes reported in figure \ref{fig:algo:repart_runtime_vs_hyperplanes}, suggesting that the overall runtime is proportional to the number of intersection checks performed. }
    %     \label{fig:repart_intersections}
    % \end{figure}
    
\begin{figure}[t!]
    \centering
    \captionsetup{width=.95\linewidth}
    \includegraphics[width=\linewidth]{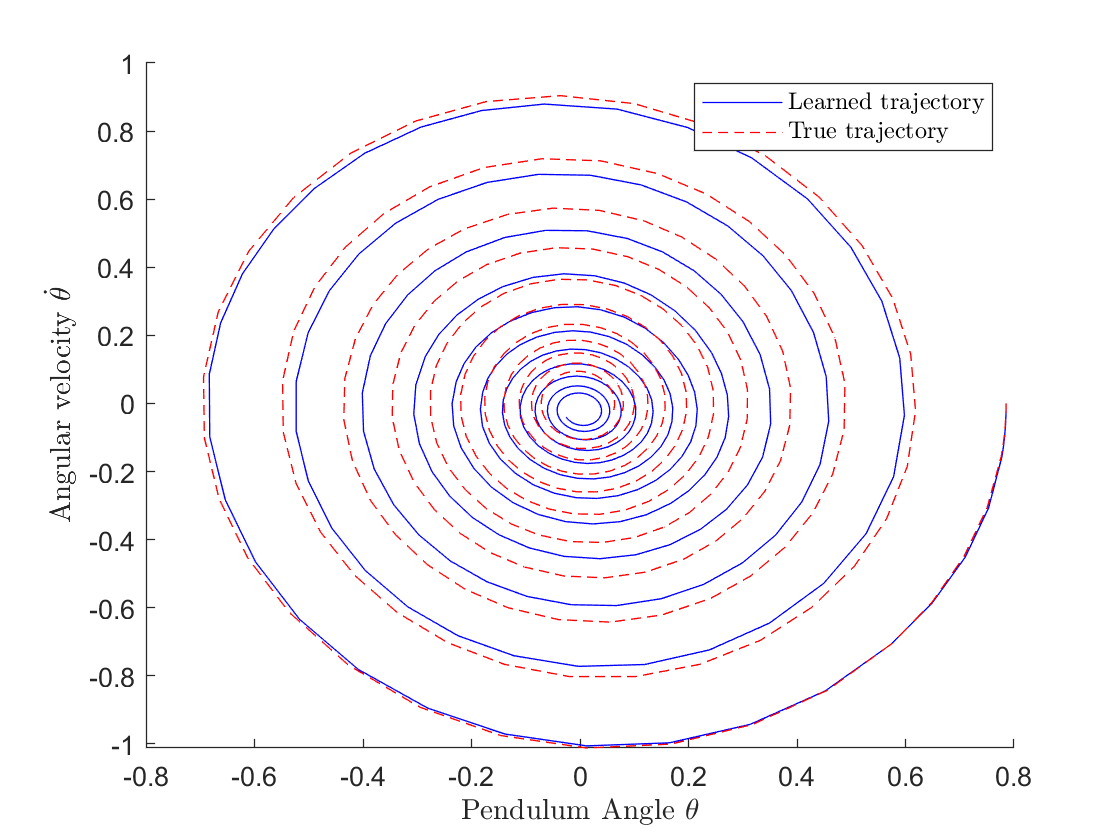}
    \caption{True and simulated trajectory using the neural network with $\theta_0 = \tfrac{\pi}{4}$. The network displays some asymmetries in its trajectory, suggesting that the learned pendulum would swing slightly higher on one side. It also appears to converge slightly off-centre of the origin. This is due to the fact that the neural network does not assume energy conservation.}
    \label{fig:learned_pendel_traj}
\end{figure}

As previously mentioned, PWA functions are widely used to represent complex dynamical systems. Neural networks are not as commonly used due to the difficulty of reasoning about their behaviour. However, it is possible train a neural network on dynamical data and then retrieve its PWA form. The algorithm is now applied to a neural network with 2 inputs, so that each of its outputs can be plotted separately as a surface. The linear regions of the network can then also be plotted in the plane. The neural network was given 2 hidden layers, with 15 and 5 neurons respectively. The network was trained on the following mapping, which describes the dynamics of a damped pendulum for $g$ = 9.81, $m$ = 1, $L$ = 5, and $d$ = 0.1.

\begin{equation}
\label{eq:pendulum}
    \ddot{\theta} = -\tfrac{g}{L} \sin{\theta} - \tfrac{d}{m}\dot{\theta} 
\end{equation}

This can be reformulated as a system of first order ordinary differential equations (ODE) where $x_1=\theta$ and $x_2=\dot{\theta}$:
\begin{equation}
\label{eq:pendulum-state-form}
    \bmat{ \dot{x}_1 \\ \dot{x}_2 } = \bmat{ x_2 \\ -\tfrac{g}{L} \sin{x_1} - \tfrac{d}{m}x_2 }
\end{equation}

\begin{figure}[t!]
        \centering
        \captionsetup{width=.9\linewidth}
        
        \begin{subfigure}{0.45\textwidth}
            \includegraphics[clip, trim=3.5cm 8.5cm 3.5cm 10cm, width=\textwidth]{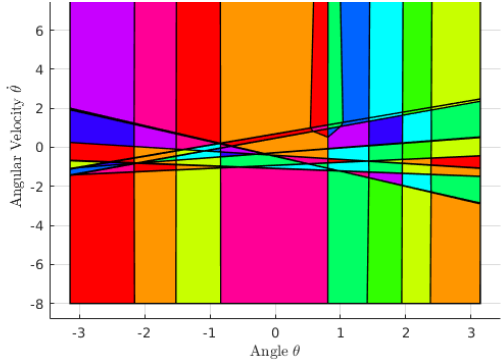}
            \caption{Linear regions (116 total) of the pendulum neural network }
            \label{fig:case2:regions}
        \end{subfigure}
        \begin{subfigure}{0.45\textwidth}
            \includegraphics[clip, trim=3.5cm 9cm 3.5cm 8.5cm, width=\textwidth]{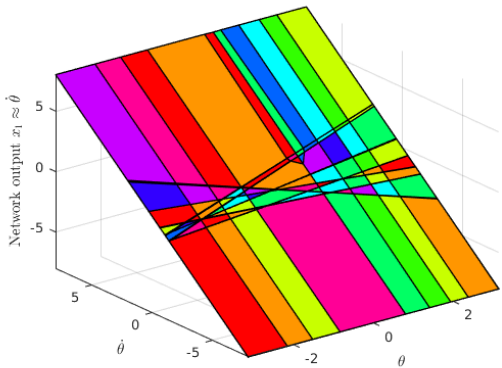}
            \caption{First output of the network: $x_1 \approx \dot{\theta}$}
            \label{fig:case2:network_out1}
        \end{subfigure}
        \begin{subfigure}{0.45\textwidth}
            \includegraphics[clip, trim=3.5cm 9cm 3.5cm 8.5cm, width=\textwidth]{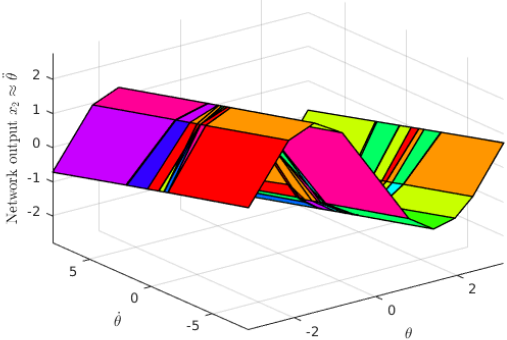}
            \caption{Second output of the network: $x_2 \approx \ddot{\theta}$}
            \label{fig:case2:network_out2}
        \end{subfigure}
        \caption{
        The complete PWA form of the pendulum dynamics neural network. The linear regions appear to have arranged themselves in patterns that support the shape of the output.}
         \label{fig:pendulum-net}
    \end{figure}

A training dataset was created by sampling $\theta$ and $\dot{\theta}$ 50000 times from the continuous uniform distribution $\mathcal{U}(-\pi, \pi)$ and the normal distribution $\mathcal{N}(0,5)$ respectively, creating a sample of states $\bm{x} = \bmat{\theta & \dot{\theta}}^T$. The corresponding $\dot{\bm{x}}$ was then found through Equation \eqref{eq:pendulum-state-form}. Then the neural network was trained on the data using the Adam (derived from ``adaptive moment estimation") optimiser with a learning rate of 0.003 for 50 epochs, finally achieving a root mean square error (RMSE) of $1.615\cdot10^{-4}$. The true and learned dynamics were then simulated using the MATLAB\supercopyright function \textit{ode45()}. Fig. \ref{fig:learned_pendel_traj} compares the two. The complete PWA form of the network is shown in Fig. \ref{fig:pendulum-net} as a pair of surface plots, along with the 116 linear regions. Interestingly, the linear regions show a concentration of horizontal boundaries around $\dot{\theta} = 0$, along with several vertical boundaries. Because the network is locally linear, the boundaries determine any changes in gradient. It is therefore likely that the concentration of horizontal boundaries serves to give the two outputs a constant slope in the $\dot{\theta}$ direction (see Fig. \ref{fig:case2:network_out1} and \ref{fig:case2:network_out2}). Likewise, the vertical boundaries form large sheets that are arranged in a sinusoidal shape that approximates Equation \eqref{eq:pendulum}. It is interesting to see such structure emerging as a result of the training process. However, there is still a bit of irregularity due to the large number of small regions between closely packed boundaries. These small regions are numerous, but highly redundant as they don't contribute significantly to the shape of the output. This can prove to be challenging when attempting to analyse the stability of such a representation, which typically involves keeping track of possible state transitions between regions, for example when using energy methods \cite{stability-pwa}. It may therefore be desirable to take steps to simplify the PWA representation either during or after the training process by merging boundaries that appear redundant, or by introducing new boundaries. This could be done by adding some kind of regularisation that forces similar connection weights for neurons in the same layer to converge together. The architecture of a network could then be simplified by merging neurons with very similar weights. Likewise, if the network is performing badly in a particular region of the state space, neurons can be split in two, introducing additional boundaries. 

\section{Conclusion}
A reasonably efficient algorithm that can obtain the PWA representation of a neural network using ReLU activation functions was presented. Results demonstrating conversions of randomly initialised neural networks with up to four dimensions and three layers were reported, the largest of which had 31835 linear regions. A parallelised version of the algorithm was able to perform this conversion in around a minute on a standard desktop computer. With more computational resources, as well as further optimisations to the algorithm, it is clear that much larger networks will be able to be converted. While this paper demonstrated how to perform the conversion for networks with fully connected layers and ReLU activations only, the approach may be generalised to any linear layer and arbitrary PWA activation functions (for example, leaky ReLU). This includes convolutional layers, normalisation layers, and networks with more complex branching architectures, which encompasses a large fraction of popular architectures in use today. The input dimension of the network appears to be a large source of complexity, limiting this approach to networks with fewer inputs. Using the method together with dimensionality reduction techniques is an approach that shows great promise for the study of complex systems that resist analysis.

\bibliographystyle{IEEEtran}
\bibliography{main}
\vskip -2\baselineskip plus -1fil
\begin{IEEEbiography}[{\includegraphics[width=1in,height=1in,clip]{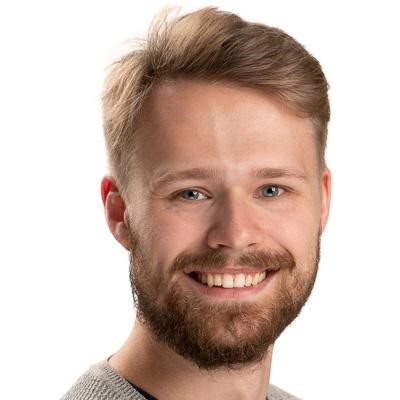}}]{Haakon Robinson} is a PhD candidate at the Norwegian University of Science and Technology (NTNU). He received a Bachelors degree in Physics in 2015 and completed a Masters degree in Cybernetics and Robotics in 2019, both at NTNU. His current work investigates the overlap between modern machine learning techniques and established methods within modelling and control, with a focus on improving the interpretability and behavioural guarantees of hybrid models that combine first principle models and data-driven components. 
\end{IEEEbiography}
\vskip -2\baselineskip plus -1fil
\begin{IEEEbiography}[{\includegraphics[width=1in,height=1.25in,clip,keepaspectratio]{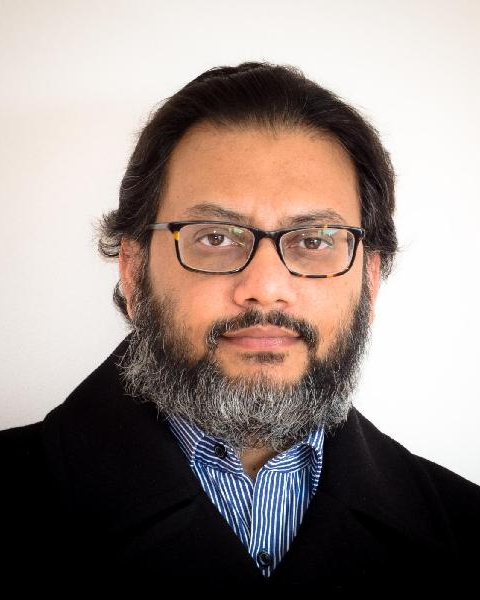}}]{Adil Rasheed} is the professor of Big Data Cybernetics in the Department of Engineering Cybernetics at the Norwegian University of Science and Technology where he is working to develop novel hybrid methods at the intersection of big data, physics driven modelling and data driven modelling in the context of real time automation and control. He also holds a part time senior scientist position in the Department of Mathematics and Cybernetics at SINTEF Digital where he led the Computational Sciences and Engineering group between 2012-2018. He holds a PhD in Multiscale Modeling of Urban Climate from the Swiss Federal Institute of Technology Lausanne. Prior to that he received his bachelors in Mechanical Engineering and a masters in Thermal and Fluids Engineering from the Indian Institute of Technology Bombay.
\end{IEEEbiography}
\vskip -2\baselineskip plus -1fil
\begin{IEEEbiography}[{\includegraphics[width=1in,height=1.25in,clip]{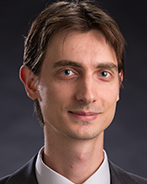}}]{Omer San} received his bachelors in aeronautical engineering from Istanbul Technical University in 2005, his masters in aerospace engineering from Old Dominion University in 2007, and his Ph.D. in engineering mechanics from Virginia Tech in 2012. He worked as a postdoc at Virginia Tech from 2012-'14, and then from 2014-'15 at the University of Notre Dame, Indiana.  
He has been an assistant professor of mechanical and aerospace engineering at Oklahoma State University, Stillwater, OK, USA, since 2015. He is a recipient of U.S. Department of Energy 2018 Early Career Research Program Award in Applied Mathematics. His field of study is centered upon the development, analysis and application of advanced computational methods in science and engineering with a particular emphasis on fluid dynamics across a variety of spatial and temporal scales. 
\end{IEEEbiography}
\end{document}